\documentclass[10pt,twocolumn,letterpaper]{article}

\usepackage{cvpr}              

\usepackage{algorithm2e}
\RestyleAlgo{ruled}
\SetKwComment{Comment}{/* }{ */}
\usepackage{multirow}
\usepackage{colortbl}
\usepackage{booktabs}
\usepackage{tabulary}
\usepackage{xcolor}
\usepackage{pifont}
\usepackage{diagbox}
\usepackage{vcell}
\usepackage{array}
\usepackage{xfrac} 
\usepackage{nicefrac}
\usepackage{amssymb}
\usepackage{marvosym}

\definecolor{cvprblue}{rgb}{0.21,0.49,0.74}
\usepackage[pagebackref,breaklinks,colorlinks,allcolors=cvprblue]{hyperref}

\def\paperID{29857} 
\def\confName{CVPR}
\def\confYear{2026}
\def\ie{\emph{i.e.}}
\def\eg{\emph{e.g.}}

\title{Unlocking Positive Transfer in Incrementally Learning Surgical Instruments: \\A Self-reflection Hierarchical Prompt Framework}

\author{
    Yu Zhu$^{1}$\qquad
    Kang Li$^{2}\textsuperscript{\Letter}$ \qquad
    Zheng Li$^{1}$ \qquad
    Pheng-Ann Heng$^{1}$ \\
    {\normalsize $^{1}$The Chinese University of Hong Kong\qquad
    $^{2}$University of Electronic Science and Technology of China}\\
    {\tt\small \{yzhu, pheng\}@cse.cuhk.edu.hk, kangli@uestc.edu.cn\textsuperscript{\Letter}, lizheng@cuhk.edu.hk}
}

\begin{document}
\maketitle
\begin{abstract}
To continuously enhance model adaptability in surgical video scene parsing, recent studies incrementally update it to progressively learn to segment an increasing number of surgical instruments over time.
However, prior works constantly overlooked the potential of \textbf{positive forward knowledge transfer}, i.e., how past knowledge could help learn new classes, and \textbf{ positive backward knowledge transfer}, i.e., how learning new classes could help refine past knowledge.
In this paper, we propose a self-reflection hierarchical prompt framework that unlocks the power of positive forward and backward knowledge transfer in class incremental segmentation, aiming to proficiently learn new instruments, improve existing skills of regular instruments, and avoid catastrophic forgetting of old instruments.
Our framework is built on a frozen, pre-trained model that adaptively appends instrument-aware prompts for new classes throughout training episodes.
To enable positive forward knowledge transfer, we organize instrument prompts into a hierarchical prompt parsing tree with the instrument-shared prompt partition as the root node, $n$-part-shared prompt partitions as intermediate nodes and instrument-distinct prompt partitions as leaf nodes, to expose the reusable historical knowledge for new classes to simplify their learning.
Conversely, to encourage positive backward knowledge transfer, we conduct self-reflection refining on existing knowledge by directed-weighted graph propagation, examining the knowledge associations recorded in the tree to improve its representativeness without causing catastrophic forgetting.
Our framework is applicable to both CNN-based models and advanced transformer-based foundation models, yielding more than 5\% and 11\% improvements over the competing methods on two public benchmarks respectively.

\end{abstract}
\section{Introduction}

\begin{figure}[t]
    \centering
    \includegraphics[width=0.455\textwidth]{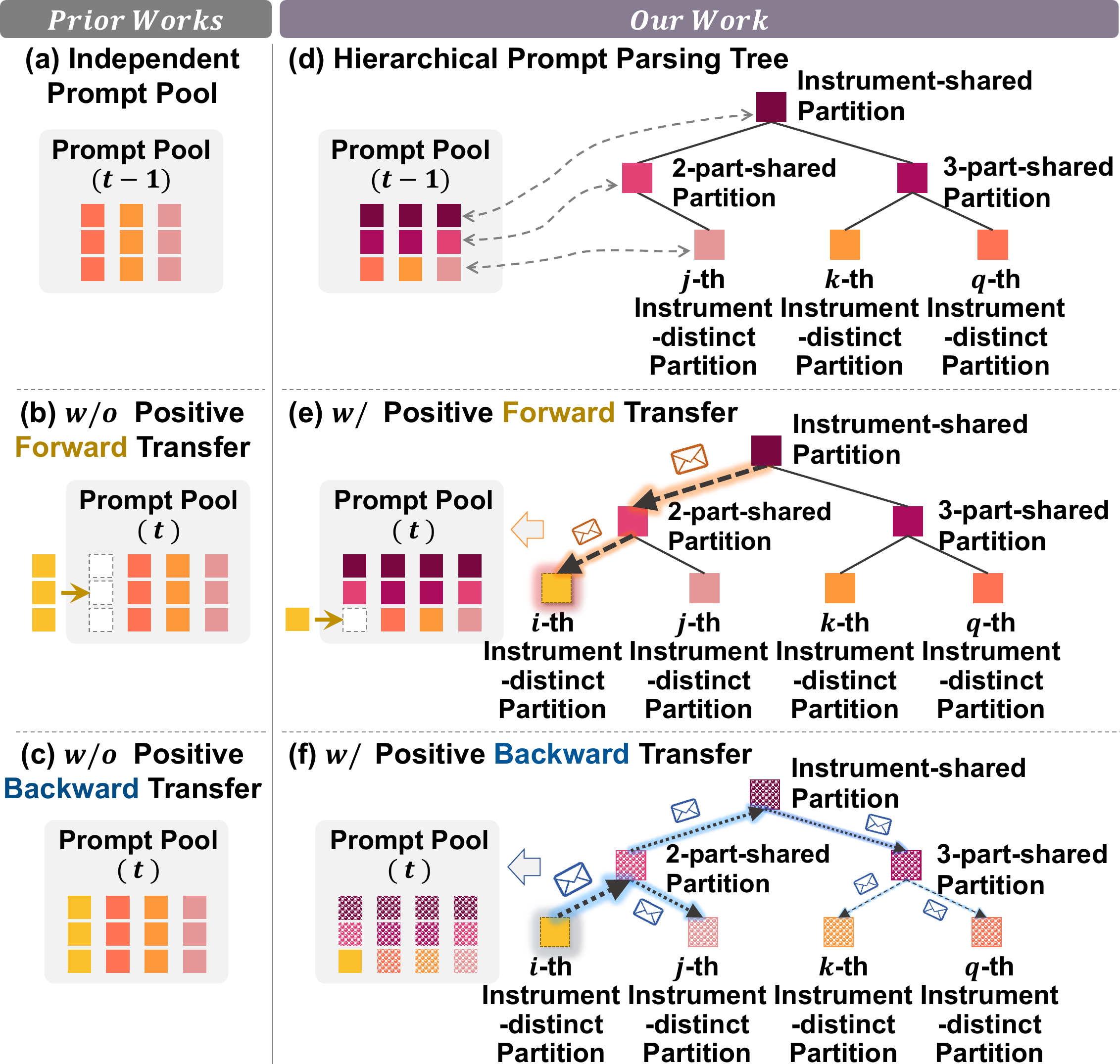}
    \caption{
    (a) Prior works learn each instrument prompt independently.
    (b,c) They overlook the power of positive forward and backward transfer between early-acquired and newly-acquired knowledge.
    (d) In contrast, our work organizes instrument prompts from general to specific as a hierarchical prompt parsing tree.
    (e) It supports positive forward transfer by helping new instrument prompts inherit profitable knowledge from the past to assist new-class learning.
    (f) It further enables positive backward transfer by letting the newly-acquired knowledge wisely refine early-acquired knowledge to improve old-class segmentation.
    }
  \label{fig:chart}
\end{figure}

In minimally invasive surgeries (\eg, cholecystectomy), surgeons normally utilize different and unfixed sets of surgical instruments depending on patient variability, intra-operative findings, potential complications, \textit{etc}, to ensure the operation is performed safely and effectively~\cite{vitiello2012emerging}.
When developing an automatic instrument segmentation model for surgical video scene parsing, 
it would be extremely time-consuming and expensive to gather training data that covers all existing instrument categories to date.
Compared to the static model training paradigm, \ie, triggering the training until all indispensable data are gathered, the dynamic model training paradigm, \ie, class incremental learning~\cite{van2022three} that iteratively exploits the available data sequentially gathered over time, to progressively cultivate the model competence for a growing number of surgical instruments, 
would be more practical to maintain an up-to-date understanding in less development time and upfront data acquisition costs.

In the general setting of class-incremental learning (CIL), the training data will arrive in a sequence over time, where each delivered dataset will bring annotations for new classes that were completely unknown in historical data.
An ideal CIL model will incrementally learn the sequentially arrived datasets to acquire all classes it has seen.

Particularly for the class-incremental segmentation of surgical instruments, the instrument categories contained in two adjacent datasets are less likely to be completely non-overlapping.
Some classes of surgical instruments (\eg, needle driver) are regularly used in routine operational actions and would appear in almost every surgical video, regardless of when it was collected (referred to as the \textit{regular} class).
However, there are also certain classes of instruments that are occasionally used in special cases (\eg, clip applier), including those that used to appear in the past but will not reappear in subsequent training episodes (\ie, the \textit{old} class), as well as those that have never appeared in the past but are newly collected to learn in current training episode (\ie, the \textit{new} class).
In each training episode, frames with irregular instruments (\eg, the new and old classes) are interspersed throughout the surgical video, blending with frames having regular instruments.
Hence, in a typical training episode of class-incremental instrument segmentation, the regular, old, and new classes concurrently exist, and a desirable CIL model should effectively acquire the ability for new instruments, advance existing abilities in regular instruments, while not catastrophically forgetting early-learned instruments, gradually growing model competence for more and more instruments.

Recent advances in continual learning increasingly emphasize the integration of foundation models by prompt tuning methods to effectively tackle various applications~\cite{li2025caprompt,wang2022learning,wang2022s}.
To elaborate, they attach the prompt parameters specific to each class into frozen foundation models, independently train prompts from scratch, and isolate them in a bank without further changes to prevent catastrophic forgetting of old classes.
However, most works underestimate the power of \textbf{positive forward knowledge transfer (PFKT)}, \ie, how past knowledge could help learn new classes, and \textbf{positive backward knowledge transfer (PBKT)}, \ie, how learning new classes could help refine past knowledge
~\cite{wang2021afec,diaz2018don,thai2022surprising}.

Undoubtedly, yielding positive forward and backward transfer would be substantially helpful to any class-incremental learning model, including the one in our case.
When incrementally learning fine-grained instrument classes, for each new instrument, one could reuse existing knowledge, such as the instrument-shared knowledge (\eg, similar long and slim shape) and $n$-part-shared knowledge (\eg, same number of compositional parts), as default prior, and only focus on acquiring the distinct features in this new instrument (\eg, special tip shapes), easing the difficulties in learning fresh classes with positive forward knowledge transfer.
Conversely, the discovery of new instrument-distinct features would trigger a whole self-introspection movement for all existing knowledge.
It examines the uniqueness of the early instrument-distinct features (\eg, black handle, L-shaped tip) and regroups those that become less distinct into common ones (\eg, black handle), refining past knowledge from self-reflection among the seen classes via positive backward knowledge transfer. 

To achieve this, we propose a self-reflection hierarchical prompt framework that unlocks the power of positive forward and backward knowledge transfer in class incremental learning to substantially assist the segmentation model in learning increasing surgical instruments over time.
Our framework gradually appends instrument-aware prompts on top of a frozen pre-trained model during training lifespan,
which instantiates new prompts for new classes and maintains existing prompts for regular and old classes to strengthen model competence.
Importantly, to encourage positive forward transfer when learning new classes, we construct a \textbf{hierarchical prompt parsing tree} that organizes all instrument-aware prompts in a three-layer tree structure, following a general-to-specific knowledge hierarchy, to help identify reusable historical knowledge for each new class.
In specific, each instrument prompt consists of (a) an instrument-shared partition (\ie, root node at the top); (b) an $n$-part-shared partition (\ie, intermediate node in the middle); and (c) an instrument-distinct partition (\ie, leaf node at the bottom).
For any new class, simply learning one instrument-distinct prompt partition would be sufficient.
By attaching this partition as a leaf node of the intermediate node with matching compositional parts, 
new-class learning can inherit instrument-shared knowledge and pertinent $n$-part-shared knowledge from the past to facilitate this process.
Based on the newly learned partition, we encourage positive backward transfer by the \textbf{self-reflection refining strategy} 
to seek necessary refinement for early-acquired knowledge.
It triggers self-reflection for each existing partition by examining the associated knowledge in the hierarchical prompt parsing tree, to discover redundancy and refine its representativeness within the context of seen classes. 
Notably, we propagate self-reflection updates by a directed-weighted graph to let it affect more for nearby nodes (with larger similarity) and fewer for distant nodes (with lesser similarity), wisely refining past knowledge without causing catastrophic forgetting.
We validated our framework on two surgical public benchmarks, which greatly outperform the competing methods in a great margin.
Our main contributions could be summarized as follows: 
\begin{itemize}
\item We propose a self-reflection hierarchical prompt framework to leverage positive forward and backward transfer to improve class-incremental instrument segmentation.

\item We propose a hierarchical prompt parsing tree to expose inheritable past knowledge for new classes during positive forward transfer.

\item We propose a self-reflection refining strategy to wisely improve past knowledge upon new-class prompts without catastrophic forgetting in positive backward transfer.

\item We validated our framework on two public surgical segmentation benchmarks in both CNN-based backbones and foundation models, where we outperformed the state-of-the-art approaches with significant gains.

\end{itemize}
\begin{figure*}[t]
  \centering
  \includegraphics[width=0.88\textwidth]{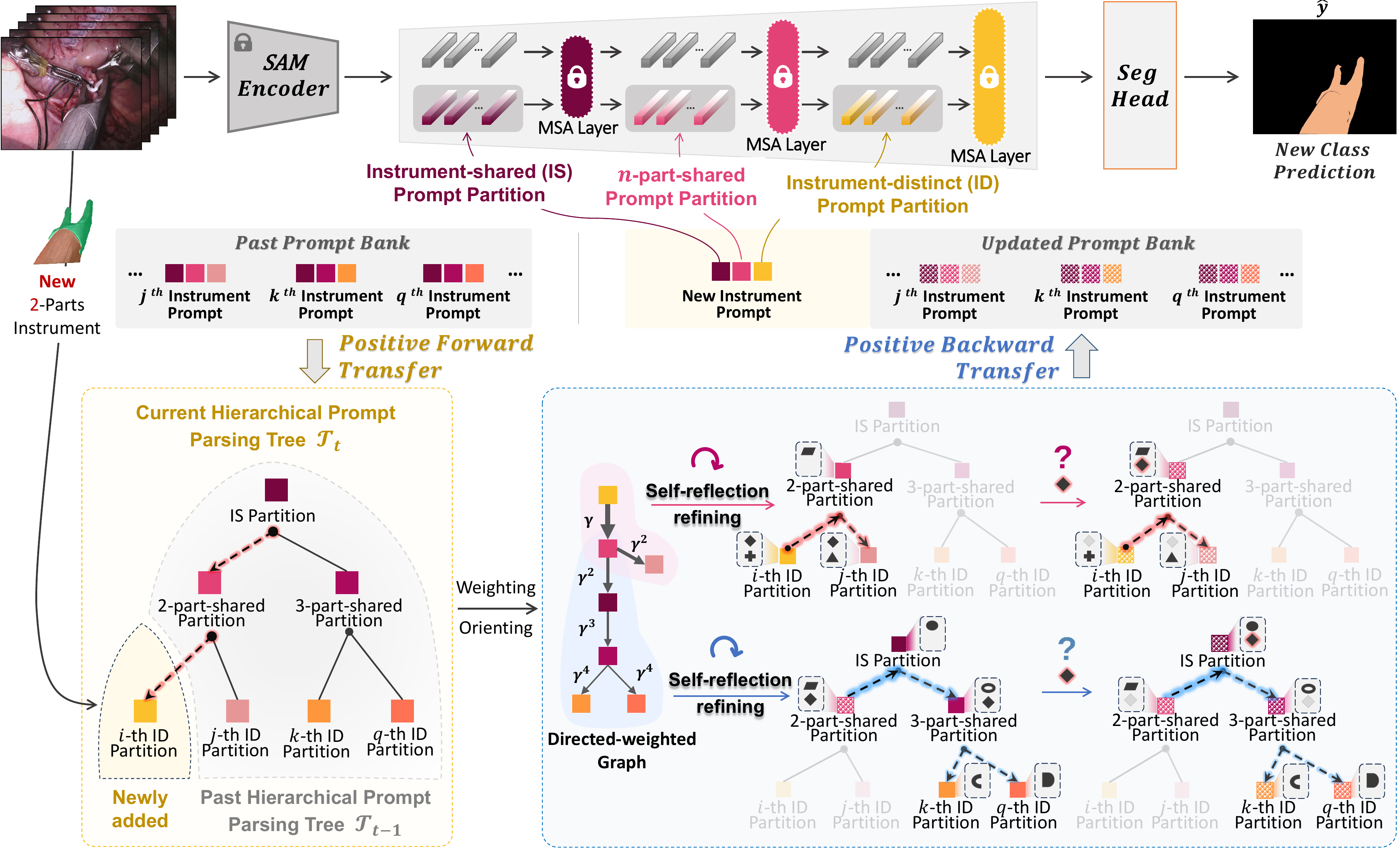}
  \caption{
  Overview of our self-reflection hierarchical prompt framework. 
  We progressively append instrument-aware prompts into a pre-trained model during the training lifespan.
  %
  For each new class, before diving into learning its specific prompts, 
  we first build a hierarchical prompt parsing tree to discover the profitable knowledge it could inherit from the past, \ie, the instrument-shared (IS) and $n$-part-shared knowledge, promoting new-class learning by positive forward knowledge transfer.
  Simply learning one instrument-distinct (ID) prompt partition would be sufficient to delineate its contour.
  Based on the newly acquired ID partition, we trigger self-reflection on existing knowledge to support positive backward knowledge transfer, wisely refining their representativeness within the context of all seen classes without causing catastrophic forgetting.
  }
  \label{fig:main-method}
\end{figure*}

\section{Related Work}
\subsection{Class Incremental Segmentation}

Most existing continual semantic segmentation works focused on the domain incremental setting~\cite{li2022domain, garg2022multi, li2024dual}.
While the class incremental setting (CIS) is more practical yet challenging in medical areas, it aims to progressively learn to segment new classes without forgetting previously learned ones.
Current CIS approaches still focus on customizing effective continual learning techniques on standard convolutional neural network (CNN) based models or simple transformer-based models~\cite{xiao2023endpoints,xu2024privacy}.
Despite their effectiveness,
their limited exploitation of powerful foundation models (\eg, SAM~\cite{kirillov2023segment}) constrains their potential for further performance improvements.
To the best of our knowledge, we are the first work that harnesses the generalization capability of SAM to promote class-incremental segmentation performance on surgical video parsing.

\subsection{Prompt Learning}
Prompt learning has been widely applied to various computer vision tasks~\cite{zhu2024memory, guo2023multiple,fischer2024prompt},
and has gained increasing attention as a key strategy for incorporating foundation models into continual learning
~\cite{li2025caprompt,qin2025pearl,wang2022learning,wang2022dualprompt}.
However, how to leverage positive forward knowledge transfer (PFKT) and positive backward knowledge transfer (PBKT) to essentially promote class incremental segmentation model performance remains under-explored.
In this paper, our work proposes a self-reflection hierarchical prompt framework to encourage previously learned knowledge to assist the learning of new classes, and encourage the newly learned knowledge to help refine previously learned knowledge in return.
Importantly, our proposed strategy is compatible with diverse backbone architectures, including CNN-based (\eg, DeepLabv3+~\cite{chen2017rethinking}) and transformer-based foundation models (\eg, SAM~\cite{kirillov2023segment}), yielding significant improvements on both architectures.

\section{Methodology}
In class-incremental segmentation, the model learns from a sequential data stream $\{\mathcal{D}_1, \mathcal{D}_2, \dots, \mathcal{D}_T\}$.
Each dataset contains the number of $N_t$ pairs of surgical video and instrument segmentation masks as $\mathcal{D}_t = \{x_j, y_j\}_{j=1}^{N_t}$, where $x_j \in \mathbb{R}^{N_v \times 3 \times H \times W}$ denotes the $j$-th video of $N_v$ image frames in the height of $H$ and width of $W$, and $y_j \in \mathbb{R}^{N_v \times H \times W}$ denotes the corresponding segmentation masks.
Let $\mathcal{C}_t =  \{c_1,c_2,\dots,c_n\}$ denote the instrument label space of $\mathcal{D}_t$, and $\mathbf{C}_{t-1} = \bigcup_{i=1}^{t-1} \mathcal{C}_i$ denote the set of all instruments seen up to the $t-1$ step.
Then, at the $t$-th training episode, it coexists a set of new instrument $\mathcal{C}_t^{new}=\{c~|~c\in \mathcal{C}_t, c \notin \mathbf{C}_{t-1} \}$ that are newly encountered in current episode, a set of regular instruments $\mathcal{C}_t^{reg} = \bigcap_{i=1}^t\mathcal{C}_i$ that regularly appeared in historical datasets, as well as a set of old instruments $\mathcal{C}_t^{old} = \mathcal{C}_t - \mathcal{C}_t^{reg} - \mathcal{C}_t^{new}$ which were previously learned but become absent in the current episode.
Due to strict privacy constraints in medical fields, only the current dataset $\mathcal{D}_t$ is accessible for training at the $t$-th episode.
An ideal CIL model aims to effectively learn new instruments, enhance the existing skills of regular instruments, and mitigate catastrophic forgetting of old instruments.

Fig.~\ref{fig:main-method} illustrates the overview of our framework.
We gradually integrate instrument-aware prompts into a pre-trained model during incremental training episodes.
For each new instrument, we first construct a hierarchical prompt parsing tree to help identify valuable knowledge that can be inherited from previous instruments, easing the difficulties of learning new classes through positive forward knowledge transfer.
Meanwhile, we leverage positive backward knowledge transfer by performing self-reflection refining on existing knowledge via directed-weighted graph propagation, examining the knowledge association within the tree to improve representativeness without causing catastrophic forgetting.

\subsection{Framework Architecture} 
During incremental learning, we gradually learn a set of trainable parameters (\ie, prompt) for seen instruments, enabling instrument-aware prompt tuning upon a well-trained backbone to accurately delineate instrument contours.
For the segmentation backbone, both CNN-based and Transformer-based encoders are applicable. 
Here, we take the well-known foundation model SAM~\cite{kirillov2023segment} as an example.
We freeze the SAM encoder $\theta_\text{Enc}$ and incorporate an adapter $\theta_a$~\cite{chen2024ma} for surgical scenario adaptation.
Instrument-aware prompts $\mathbf{P}^c$ are attached to different depths of the mask decoder $\theta_{\text{Dec}}$ from shallow to deep, and the original class-agnostic segmentation head is replaced by multiple instrument-aware segmentation heads $\theta_{\text{Seg}}^c$. 
$\theta_a$ and $\theta_{\text{Dec}}$ are trained only at the first step and frozen thereafter, while $\mathbf{P}^c$ and $\theta_{\text{Seg}}^c$ are incrementally appended and updated.
For new classes, we initialize new corresponding instrument-aware prompts and segmentation heads, while for old and regular classes, the segmentation heads are frozen and only the prompts are updated.

\subsection{Hierarchical Prompt Parsing Tree}
Instead of letting these instrument prompts be independent of each other as prior works did~\cite{wang2022dualprompt,wang2022learning}, we organize instrument-aware prompts as a hierarchical prompt parsing tree to explore the relation among instrument prompts.
The prompt parsing tree has three layers: the root node is the instrument-shared partition $\text{P}_\text{IS}$; it branches into intermediate nodes $\{\text{P}_\text{n-p}^{i}\}_{i=1}^n$, each would be shared across instruments with $n$ parts; each intermediate node further connects to leaf nodes $\{\text{P}_\text{ID}^{c}\}_{c \in \mathbf{C}_t}$, where each $\text{P}_\text{ID}^{c}$ is distinctive to the $c$-th instrument.
In this regard, when new instruments appear, we can effectively identify reusable knowledge from previously learned knowledge, thereby facilitating positive knowledge transfer and promoting the learning of new classes.
In the initial step $t=1$, for $c$-th instrument with $i$-parts, its prompt $\mathbf{P}^c$ would consist of three prompt parsing tree nodes: (1) root node $\text{P}_\text{IS}$; (2) intermediate node $\text{P}_\text{n-p}^{i}$; and (3) leaf node $\text{P}_\text{ID}^{c}$.
Prompt partitions are inserted into different decoder layers, from shallow (\eg, the $l_1$-th layer) to deep (\eg, the $l_3$-th layer).
In the $l$-th decoder layer, given the input feature $F^l$, prompt tuning is applied during the cross-attention mechanism. 
The prompt partition $\text{P} \in \mathbb{R}^{u \times b}$ serves as the query, where $u$ represents the number of prompt tokens and $b$ denotes the embedding dimension. 
The position-embedded input feature $f_{pe}(F^l)$ is used as the key, and $F^l$ as the value, computed as: 
\begin{equation}
    h_{atten}^{(l)} (F^l,\text{P}) = f_{atten}(\text{P}, f_{pe}(F^l), F^l).
\end{equation}
Accordingly, the hierarchical prompt tuning process $f_{hpt}$ with all three prompt partitions in $\mathbf{P}^c$ could be denoted as
\begin{equation}
    \begin{aligned}
    &f_{hpt} (F^{l}, [\text{P}_\text{IS};\text{P}_\text{n-p}^{n};\text{P}_\text{ID}^{c}])\\ &=h_{atten}^{(l_3)}(h_{atten}^{(l_2)}(h_{atten}^{(l_1)}(F^{l},\text{P}_\text{IS}), \text{P}_\text{n-p}^{n}), \text{P}_\text{ID}^{c}).
\end{aligned}
\end{equation}
We then feed the output of the final decoder layer into the segmentation head $\theta_\text{Seg}^c$ for producing the final segmentation mask.
The objective function in the first training episode could be further elaborated as:
\begin{equation}
\label{step1 obj func}
\min _{\substack{\theta_a,\theta_{\mathrm{Seg}}^c, \theta_{\mathrm{Dec}},\\ [\text{P}_\text{IS};\text{P}_\text{n-p}^{n};\text{P}_\text{ID}^{c}]}\!\!\!\!\!\!\!\!\!\!} \mathcal{L}_{\mathrm{ce}}\left(\theta_{\text{Seg}}^c\left(\theta_{\text{Dec}}\left(f_{\text {Enc}}\left(x, \theta_a\right), [\text{P}_\text{IS};\text{P}_\text{n-p}^{n};\text{P}_\text{ID}^{c}]\right)\right), y^c\right),
\end{equation}
where $(x,y^c)\in D_1$ and $y^c$ is the mask for instrument $c$.
$\text{P}_\text{IS}$ is optimized for all instruments, $\text{P}_\text{n-p}^{n}$ is updated only when the instrument consists of $n$ parts, and $\text{P}_\text{ID}^{c}$ is specifically optimized for instrument $c$, ensuring that each prompt partition learns relevant knowledge under proper supervision. 

After optimization, the trained prompt partition is organized into a hierarchical prompt parsing tree, represented as an undirected graph $\mathcal{T}^{1} = (\mathcal{V}^1, \mathcal{E}^1)$, and forwarded to the next training episode.
We denote its vertex set as $\mathcal{V}^1 = \{\text{P}_\text{IS}\} \cup \{\text{P}_\text{n-p}^{i}\}_{i=1}^{n}\cup\{\text{P}_\text{ID}^{c}\}_{c=1}^{N_1}$, where $N_1=|\mathbf{C_1}|$ and $n$ is the maximum number of composition parts (set to 3 following common practice in surgical videos).
We further denote its edge set as 
$\mathcal{E}^1=\{(\text{P}_\text{IS}, \text{P}_\text{n-p}^{i})|~i\in [1, n]\} \cup \{(\text{P}_\text{n-p}^{i}, \text{P}_\text{ID}^{j})|~i\in [1, n], j \in [1,N_1]\}$
, where each element $(v,u)$ indicates an edge connecting the node $v$ and the node $u$ in $\mathcal{T}^{1}$.

In subsequent training steps $t \geq 2$, once a newly-seen class $c$ composed of $k$ parts is encountered, relevant prior knowledge can be efficiently retrieved from the previous hierarchical prompt parsing tree $\mathcal{T}^{t-1}$.
Specifically, the retrieved knowledge includes the root node $\text{P}_\text{IS}$, which is shared across all instruments, and the intermediate node $\text{P}_\text{n-p}^{k}$, which captures shared semantics among instruments composed of $k$ parts.
%
By simply inserting a newly initialized instrument-specific prompt $\text{P}_\text{ID}^{c}$ under $\text{P}_\text{n-p}^k$, the prior knowledge can be effectively inherited.
The hierarchical prompt parsing tree then became $\mathcal{T}^t = \{\mathcal{V}^{t-1} \cup \{\text{P}_\text{ID}^{c}\}, \mathcal{E}^{t-1} \cup \{(\text{P}_\text{n-p}^k, \text{P}_\text{ID}^{c})\}$. 
We then optimize the newly introduced prompt partition $\text{P}_\text{ID}^{c}$ as well as the related segmentation head $\theta_{\text{Seg}}^c$ by the following objective:
\begin{equation}
\label{step2_node_seg}
    \min _{\substack{\theta_{\text{Seg}}^c, \text{P}_\text{ID}^{c}}} \mathcal{L}_{\text{ce}}\left(\theta_{\text {Seg}}^{c}\left(\theta_{\text {Dec}}\left(f_{\text{Enc}}\left(x, \theta_a\right), [\text{P}_\text{IS};\text{P}_\text{n-p}^{n};\text{P}_\text{ID}^{c}]\right)\right), y^c\right).
\end{equation}
Consequently, the model enables positive forward transfer and acquires the distinct knowledge $\text{P}_\text{ID}^c$ of class $c$.

\subsection{Self-reflection Refining Strategy}
\begin{algorithm}[t]
\caption{Training Procedures} 
\label{algorithm}
{\bf Output:} 
$\theta_{a}$, $\theta_\text{Dec}$, $\{\theta_\text{Seg}^c\}_{c \in \mathbf{C}_t}$, and $\{\mathbf{P}^c\}_{c \in \mathbf{C}_t}$. 
\\
\While{incrementally learning from $t=1$ to $T$}{
    \eIf{$t == 1$}{
        Load and freeze pre-trained encoder as $\theta_\text{Enc}$.\\
        Optimize  $\{\theta_\text{Seg}^c\}_{c \in \mathbf{C}_1}$ , $\{\mathbf{P}^c\}_{c \in \mathbf{C}_1}$ , $\theta_\text{Dec}$, and $\theta_{a}$ with $D_1$ by Eq.~\ref{step1 obj func} and construct $\mathcal{T}^1$.\\
    }{
        \For{$c \in  \mathcal{C}_t^{new}$}{
        Initialize new-class node $\text{P}_\text{ID}^{c}$.
        Insert $\text{P}_\text{ID}^{c}$ into $\mathcal{T}^{t-1}$ to obtain $\mathcal{T}^{t}$.\\
        Freeze $\theta_\text{Enc}$, $\theta_\text{Dec}$, $\theta_{a}$, and $\{\theta_\text{Seg}^c\}_{c \in  \mathcal{C}_t^{old}}$, and train $\text{P}_\text{ID}^{c}$ and $\theta_\text{Seg}^c$ with $D_t$ by Eq.~\ref{step2_node_seg}.\\
        Obtain $\mathcal{G}_c^t= (\mathcal{V}^t_c, \tilde{\mathcal{E}^t_c})$ by orienting and weighting the current $\mathcal{T}^t$.\\
        Conducting self-reflection via Eq.~\ref{digobject funtion} to derive the refined $\tilde{\mathcal{G}_c^t} = (\tilde{\mathcal{V}^t_c}, \tilde{\mathcal{E}^t_c})$.\\
        Update $\mathcal{T}^t$ by $\tilde{\mathcal{V}^t_c}$.\\}
    }
    Pass $\theta_\text{Enc}$, $\theta_{a}$, $\theta_\text{Dec}$, $\{\theta_\text{Seg}^c\}_{c \in \mathbf{C}_t}$, $\{\mathbf{P}^c\}_{c \in \mathbf{C}_t}$ , and $\mathcal{T}^t$ to the $t+1$ step.
}
{\bf Return} $\theta_{a}$, $\theta_\text{Dec}$, $\{\theta_\text{Seg}^c\}_{c \in \mathbf{C}_t}$, $\{\mathbf{P}^c\}_{c \in \mathbf{C}_t}$. 
\end{algorithm}
\begin{figure*}[t]
  \centering
  \includegraphics[width=1\textwidth]{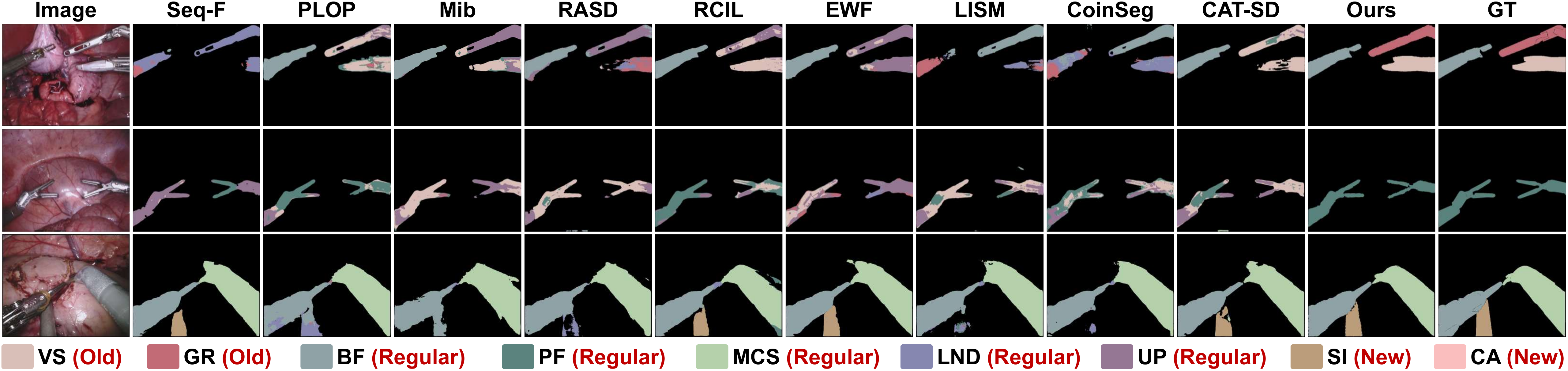}
  \caption{Visual comparisons of our approaches and highly competitive approaches. More comparison results are in the supplementary.
  }
  \label{vis}
\end{figure*}
\begin{table*}[t]
\centering
\caption{
Porcine robot-assisted nephrectomy instrument segmentation results for the last episode evaluated by IoU and all episodes assessed by BWT and FWT. For all metrics, the higher the better. We highlight the best and second-best results in \textbf{bold} and \underline{underline}, respectively.
}
\label{tab1porcine}
\resizebox{\linewidth}{!}{ 
\begin{tabular}{l|cc|ccccc|cc|c|c|c|c|c|c}
\toprule[1pt]
\multicolumn{1}{c|}{\multirow{2}{*}{Method}} & \multicolumn{2}{c|}{Old Class} & \multicolumn{5}{c|}{Regular Class} & \multicolumn{2}{c|}{New Class} & \multirow{2}{*}{\begin{tabular}[c]{@{}c@{}}Old \\ Class\end{tabular}} \rule{0pt}{2ex}& \multirow{2}{*}{\begin{tabular}[c]{@{}c@{}}Regular\\ Class\end{tabular}} & \multirow{2}{*}{\begin{tabular}[c]{@{}c@{}}New\\ Class\end{tabular}} & \multirow{2}{*}{All} & \multirow{2}{*}{BWT}& \multirow{2}{*}{FWT} \\ \cline{2-10}
\multicolumn{1}{c|}{}  & VS\rule{0pt}{2ex}  & GR & PF & BF & UP & LND & MCS & SI & CA &  &  &  & & & \\ \midrule
Ind-T ($t=1$) & 75.12 & 62.50 & 68.27 & 66.22 & 64.65 & 71.89 & 70.91 & - & - & 68.81 & 68.39 & - & 68.51& -& - \\
Ind-T ($t=2$) & - & - & 69.67 &	67.67 &	66.44 &	70.63 &	70.80 &	32.98 &	9.18 & - & 69.04 & 21.08 & 55.34 & -& -\\
Joint-T & 83.46 & 70.91 & 79.79 & 71.45 & 71.88 & 71.08 & 75.04 & 35.94 & 8.90 & 77.18 & 73.85 & 22.42 & 63.16 & -& -\\ \midrule
Seq-F & 4.90 & 2.97 & 48.73 & 52.23 & 44.81 & 46.73 & 37.47 & 9.45 & 1.29 & 3.93 & 45.99 & 5.37 & 27.62 & -8.92 & 0.14\\ 
SI & 6.51 & 2.65 & 50.26 & 58.91 & 29.58 & 51.29 & 42.02 & 11.48 & 1.82 & 4.58 & 46.41 & 6.65 & 28.28 & -8.44 & 1.42\\
LWF & 7.64 & 3.99 & 49.86 & 49.48 & 36.97 & 59.13 & 23.81 & 4.97 & 0.36 & 5.82 & 43.85 & 2.67 & 26.25 &-9.92&-2.56\\
ILT & 7.61 & 1.71 & 33.57 & 55.31 & 30.44 & 59.02 & 51.86 & 0.61 & 1.67 & 4.66 & 46.04 & 1.14 & 26.87 &-8.68&-4.09\\
PLOP & 11.14 & 3.42 & 37.50 & 57.63 & 31.89 & 61.91 & 49.57 & 5.39 & 0.27 & 7.28 & 47.70 & 2.83 & 28.75 &-6.75 	&-2.40 
\\
MiB & 13.01 & 10.85 & 38.76 & 55.01 & 36.48 & 36.21 & 45.07 & 8.98 & 2.91 & 11.93 & 42.30 & 5.95 & 27.47 &-9.28 	&0.72 
\\
RASD & 8.27 & 3.23 & 52.11 & 63.46 & 45.74 & 63.78 & 42.36 & 11.60 & 3.74 & 5.75 & 53.49 & 7.67 & 32.70 &-3.05 	&2.44 
\\
RCIL & 12.19 & 2.70 & \underline{54.89} & 61.51 & 53.98 & 64.93 & 35.06 & 4.11 & 2.74 & 7.45 & 54.07 & 3.42 & 32.46 &-2.15 	&-1.80 
\\
EWF & 14.03 & 13.58 & 53.39 & 52.96 & 54.66 & 54.33 & 50.12 & 11.24 & 1.83 & 13.81 & 53.09 & 6.54 & 34.02 &-1.03 	&1.31 
\\
LISM & 22.24 & 5.83 & 46.66 & 56.42 & 51.16 & 51.97 & 55.21 & 15.51 & 5.50 & 14.03 & 52.28 & 10.50 & 34.50 &-1.55 	&5.27 
\\
CoinSeg & 12.78 & 9.43 & 44.81 & \underline{64.72} & 49.68 & 68.03 & 51.88 & 4.75 & 1.11 & 11.10 & 55.82 & 2.93 & 34.13 &0.14 	&-2.30 
\\
CAT-SD & 22.88 & 16.65 & 43.95 & 64.50 & 36.69 & 57.86 & 51.61 & 15.91 & 5.80 & 19.76 & 50.92 & 10.85 & 35.09 &-0.88 	&\underline{5.62}
\\ \midrule
Ours (DeepLabv3+) & \underline{27.28}	& \underline{17.64}	& 54.81	& 63.98	& \underline{55.13}	& \underline{68.41}	&	\underline{56.63} &	\underline{16.43} & \underline{	7.96}	&	\underline{22.46}	& \underline{59.79}	& \underline{12.19}	& \underline{40.92}&\textbf{6.22} 	&\textbf{6.96}
\\
Ours (SAM)& \textbf{75.46} & \textbf{61.58} & \textbf{73.48} & \textbf{65.67} & \textbf{69.97} & \textbf{72.03} & \textbf{72.19} & \textbf{35.28} & \textbf{9.32} & \textbf{68.52} & \textbf{70.67} & \textbf{22.30} & \textbf{59.44}& \underline{1.55}& 	1.22 
\\ \bottomrule[1pt]
\end{tabular}
}
\end{table*}
Based on the current hierarchical prompt parsing tree $\mathcal{T}^{t}$ with the newly appended instrument-distinct partition $\text{P}_\text{ID}^{c}$, we then perform positive backward transfer to refine the early-acquired knowledge by wisely adjusting existing partitions in $\mathcal{T}^{t}$.
To achieve that, we construct a directed and weighted graph $\mathcal{G}^{t}_c$ with the same vertex set $\mathcal{V}^t$ as $\mathcal{T}^{t}$, but assigning direction and weights to its edge set into $\tilde{\mathcal{E}^t_c}$.
Specifically, we set $\text{P}_\text{ID}^{c}$ as the root node and all edges in $\mathcal{E}^t$ follow an outward flow from the root node $\text{P}_\text{ID}^{c}$, connecting nodes in order of increasing distance from $\text{P}_\text{ID}^{c}$ and represented as:
\begin{equation}
\label{directed}
\begin{aligned}
\tilde{\mathcal{E}^t_c} = \!\!\bigcup_{v_i, v_j \in \mathcal{V}^t}\!\! \left\{ \left[v_i, v_j\right] \mid \mathrm{d}\left(v_i, \text{P}_\text{ID}^{c}\right) < \mathrm{d}\left(v_j, \text{P}_\text{ID}^{c}\right) \right\},
\end{aligned}
\end{equation}
where $\mathrm{d}(\cdot)$ denotes the length of the unique path connecting them in the current graph $\mathcal{G}^{t}_c$ (\ie, the number of edges along that path). 
$\left[\cdot, \cdot\right]$ represents a directed edge where $\left[v_i, v_j\right]\not\equiv\left[v_j, v_i\right]$.
We further assign a weight \( W_{[v_i,v_j]} = \gamma^{\mathrm{d}(v_j, \text{P}_\text{ID}^{c})} \) to each edge in \( \mathcal{G}^{t}_c \), where \( \gamma \in (0,1) \) is a decay factor.
The decay factor ensures that self-reflection primarily focuses on relevant and recent available classes, thereby avoiding catastrophic forgetting by maintaining the stability of knowledge related to old classes or less relevant classes. 
%
%
Accordingly, the adjacency matrix $\mathcal{A}_c^t \in\mathbb{R}^{|\mathcal{V}^t| \times |\mathcal{V}^t|}$ of $\mathcal{G}^{t}_c$ is defined as $\mathcal{A}^{t}_{c}|_{i,j} = \mathbb{I}_{[v_i, v_j] \in \tilde{\mathcal{E}^t_c}} \cdot W_{[v_i,v_j]}$, where $\mathbb{I}$ is the indicator function.
We further employ a directed graph network $f_{dign}$~\cite{tong2020digraph} to achieve self-reflection between the new prompt partition and the existing prompt partitions by directed graph convolution.
Specifically, we first calculate the transition matrix $\mathbf{T}^c$ by:
\begin{equation}
\label{dicov}
\begin{aligned}
\mathbf{T}^c =\left(1 - \alpha\right)\mathbf{D}^{-1}\mathcal{A}_c^t + \frac{\alpha}{n} \mathbf{1}^{|\mathcal{V}^t| \times |\mathcal{V}^t|},\\
\end{aligned}
\end{equation}
where $\mathbf{D}(i, i) = \sum_j \mathcal{A}_c^t(i, j)$ is the degree matrix, and $\alpha\in (0, 1)$ is a hyperparameter known as the teleport probability.
The term $\frac{\alpha}{n} \mathbf{1}^{n \times n}$ is a teleportation matrix~\cite{page1999pagerank} that ensures $\mathbf{T}^c$ is irreducible and aperiodic and thereby guarantees the existence of a unique left eigenvector $\pi^c$ with eigenvalue 1~\cite{barker1975algebraic}.
The output of the directed graph convolution is:
\begin{equation}
\label{dicov}
\begin{aligned}
\tilde{\mathcal{G}^{t}_c}=&(\tilde{\mathcal{V}^t_c}, \tilde{\mathcal{E}^t_c})=f_{dign}(\mathcal{V}^t, \tilde{\mathcal{E}^t_c})\\
=&\frac{1}{2} \left( 
\Pi^{\frac{1}{2}} \mathbf{T}^c \Pi^{-\frac{1}{2}} 
+ 
\Pi^{-\frac{1}{2}} \mathbf{T}^{c\top} \Pi^{\frac{1}{2}} 
\right) \mathbf{F}\left(\mathcal{V}^t\right)\Theta,
\end{aligned}
\end{equation}
where $\Pi = \frac{1}{\|\pi^c\|_1} \, \mathrm{Diag}(\pi^c)$, $\mathbf{F}(\cdot)$ is the flatten operation, and $\Theta$ is the learnable parameter.
We optimize $f_{dign}$ to produce $\tilde{\mathcal{G}^{t}_c}$ that performs well across all currently available classes as follows:
\begin{equation}
\label{digobject funtion}
\begin{aligned}
\min_{f_{dign} }\mathcal{L}_{\text{ce}}\left( \theta_\text{Seg}^c\left(\theta_\text{Dec}\left( f_\text{Enc}\left(x,\theta_a\right); f_{dign}\left( \mathcal{G}_c^t \right) \right)\right), y^c \right).
\end{aligned}
\end{equation}
After optimization, the refined prompt partition tree is derived from \( \tilde{\mathcal{G}}^t_c \) by discarding edge weights and directions.
The overall training scheme is presented in Algorithm 1. 
%
%

\section{Experiments}
\begin{table*}[t]
\centering
\caption{
Laparoscopic cholecystectomy instrument segmentation results for the last episode evaluated by IoU and all episodes assessed by BWT and FWT. For all metrics, the higher the better. We highlight the best and second-best results in \textbf{bold} and \underline{underline}, respectively.
}
\label{tab2}
\resizebox{0.85\linewidth}{!}{
\begin{tabular}{l|c|c|ccccc|c|c|c|c}
\toprule[1pt]
\multicolumn{1}{c|}{\multirow{2}{*}{Method}} & Old Class & Regular Class & \multicolumn{5}{c|}{New Class}  & \multirow{2}{*}{\begin{tabular}[c]{@{}c@{}}New\\Class\end{tabular}} & \multirow{2}{*}{All}& \multirow{2}{*}{BWT}& \multirow{2}{*}{FWT} \\ \cline{2-8}
\multicolumn{1}{c|}{} & LH\rule{0pt}{2ex} & Gr & Cl & Sc & Ir & SB & Bi &  & & & \\ \midrule
Ind-T ($t=1$) & 76.04 & 74.11 & - & - & - & - & - & - & 75.08 & - & - \\
Ind-T ($t=2$) & - & 79.40 & 69.34 & 47.02 & 39.05 & 63.91 & 59.56 & 55.78 & 59.71 & - & -\\
Joint-T & 79.54 & 80.88 & 71.07 & 49.03 & 39.64 & 67.65	&57.87&	57.05&	63.67& - & -\\  
\midrule
Seq-F & 39.07 & 53.67 & 62.59 & 49.61 & 32.18 & 47.14 & 43.50 & 47.01 & 46.82 &-12.52 &	-1.00 
\\
SI & 41.65 & 54.30 & 63.15 & 49.70 & 31.86 & 50.75 & 47.60 & 48.61 & 48.43 &-10.92 &	0.60 
\\
LWF & 36.29 & 54.70 & 65.46 & 42.98 & 35.16 & 45.23 & 51.94 & 48.15 & 47.39 & -13.40 &	0.14 
\\
ILT & 39.43 & 52.52 & 55.51 & 48.77 & 30.32 & 54.39 & 52.29 & 48.25 & 47.60 &-12.92 &	0.24 
\\
PLOP & 46.29 & 58.98 & 65.34 & 33.41 & 33.12 & 50.80 & 54.78 & 47.49 & 48.96 & -6.26 &	-0.52 
\\
MiB & 42.29 & 47.79 & 65.31 & 38.48 & 33.25 & 55.48 & 50.48 & 48.60 & 47.58 & -13.85 &	0.59 
\\
RASD & 55.53 & 51.54 & 50.40 & 44.93 & 32.29 & 51.04 & 52.23 & 46.18 & 48.28 & -5.36 &	-1.83 
\\
RCIL & 43.97&	55.89&	61.68&	42.53&	29.10&	55.39&	52.61&	48.26&	48.74 & -8.96 &	0.25 
\\
EWF & 43.38 & 53.96 & 65.89 & 44.70 & 31.96 & 58.27 & 50.73 & 50.31 & 49.84 & -10.23 &	\underline{2.30}
\\
LISM & 45.80 & 51.06 & 61.06 & 45.20 & 29.27 & 50.72 & 50.47 & 47.34 & 47.65 & -10.47 &	-0.67 
\\
CoinSeg & 54.45 & 48.30 & 57.09 & 42.27 & 35.65 & 53.34 & 54.72 & 48.61 & 49.40 & -7.52 &	0.60 
\\
CAT-SD & 55.14 & 62.39 & \underline{66.05} & 43.98 & 34.66 & 47.35 & 52.63 & 48.93 & 51.74 & -0.13 &	0.92 
\\ \midrule
Ours (DeepLabv3+)& \underline{58.08} & \underline{66.18} & 65.73 & \underline{50.17} & \underline{35.88} & \underline{61.38} & \underline{55.41} & \underline{53.71} & \underline{56.12} & \textbf{3.24} &	\textbf{5.70} 
\\ 
Ours (SAM) & \textbf{74.90} & \textbf{80.85} & \textbf{71.04} & \textbf{50.29} & \textbf{38.93} & \textbf{64.98} & \textbf{57.10} & \textbf{56.47} & \textbf{62.58}&\underline{2.80} &	0.69 
\\ \bottomrule[1pt]
\end{tabular}
}

\end{table*}

\subsection{Dataset and Experiment Settings}
\textbf{Porcine Robot-assisted Nephrectomy}\quad
We utilize two datasets, EndoVis 2017 ($t=1$)~\cite{allan20192017} and EndoVis 2018 ($t=2$)~\cite{allan20202018, gonzalez2020isinet}, to construct a continuous data stream in order for our experiments. 
The vessel sealer (VS) and grasping retractor (GR) appear only in EndoVis 2017 and are therefore considered old classes in our setup.
The prograsp forceps (PF), bipolar forceps (BF), ultrasound probe (UP), large needle driver (LND), and monopolar curved scissors (MCS) are regular classes, as they are present in both EndoVis 2017 and EndoVis 2018.
The suction instrument (SI) and clip applier (CA) appear only in EndoVis 2018, which are treated as new classes.

\textbf{Laparoscopic Cholecystectomy}\quad
A continuous data stream is built by sequentially combining two datasets, CholecSeg8k ($t$=1)~\cite{hong2020cholecseg8k} and M2CAI-Seg ($t$=2)~\cite{maqbool2020m2caiseg}.
The L-hook electrocautery (LH) from CholecSeg8k is an old class, while the grasping retractor (GR) is the regular class.
The Clipper (Cl), Scissors (Sc), Irrigator (Ir), Specimen bag (SB), and Bipolar (Bi) are new classes.

\begin{table}[t]
\centering
\caption{
Ablation study of our key components.
}
\resizebox{\linewidth}{!}{
\begin{tabular}{c|c|c|c|c|c|c|c}
\toprule[1pt]
\multicolumn{1}{c|}{DLv3+} & \multicolumn{1}{c|}{SAM} & \multicolumn{1}{c|}{HPPT} & \multicolumn{1}{c|}{SRS} & \multicolumn{1}{c|}{\begin{tabular}[c]{@{}c@{}}Old \\ Classes\end{tabular}} & \multicolumn{1}{c|}{\begin{tabular}[c]{@{}c@{}}Regular \\ Classes\end{tabular}} & \multicolumn{1}{c|}{\begin{tabular}[c]{@{}c@{}}New \\ Classes\end{tabular}} & \multicolumn{1}{c}{All}  \\ \midrule
 & \checkmark &  &  & 21.42 & 69.11 & 22.24 & 48.15 \\
 & \checkmark & \checkmark & & 65.29 & 69.23 & 22.16	 & 57.89  \\		
 & \checkmark & \checkmark & \checkmark & 68.52 & 70.67 & 22.30 & 59.44 \\
 \checkmark &  &  &  & 3.93 & 45.99 & 5.37 & 27.62 \\
 \checkmark &  & \checkmark & \checkmark & 22.46 & 59.79 & 12.19 & 40.92 \\ \bottomrule[1pt]
\end{tabular}
}

\label{tab3ablation}
\end{table}

\textbf{Implementation Details}\quad
We adopt the SAM ViT-H~\cite{kirillov2023segment} version as our encoder $\theta_\text{Enc}$. 
To better evaluate model performance in class-incremental learning (CIL), we adopt the metrics of Backward Transfer (BWT) and Forward Transfer (FWT)~\cite{apolinario2025code,diaz2018don,lopez2017gradient}. 
In the CIL setting, BWT measures how learning new classes affects performance on old and regular classes. 
FWT shows how prior learning supports new classes. 
Their detailed equations and further implementation details are provided in the supplementary.

\subsection{Compared Methods} The straightforward approach is to sequentially fine-tune the previous model on each episode (Seq-F).
We also compare our method with other state-of-the-art methods, including SI~\cite{zenke2017continual}, LWF~\cite{li2017learning}, ILT~\cite{michieli2019incremental}, PLOP~\cite{douillard2021plop}, MiB~\cite{cermelli2020modeling}, RASD~\cite{michieli2021continual}, RCIL~\cite{zhang2022representation}, EWF~\cite{xiao2023endpoints}, LISM~\cite{liu2022learning}, CoinSeg~\cite{zhang2023coinseg} and CAT-SD~\cite{xu2024privacy}.
Additionally, we train a model individually for each incremental step (Ind-T), as well as jointly train a model (Joint-T) where all classes are available from all incremental steps.
For fair and efficient comparison, both Ind-T and Joint-T are trained using the powerful SAM~\cite{kirillov2023segment} as the backbone.
%
Joint-T represents the upper bound since it is trained with all classes at once.
For fair comparison with other state-of-the-art CIL methods, we also implement our approach using DeepLabv3+~\cite{chen2017rethinking} as the backbone.
\subsection{Experiment Results}
We report the performance of the instrument segmentation in porcine robot-assisted nephrectomy in Table~\ref{tab1porcine}. 
At the last learning episode, our method significantly outperforms all state-of-the-art methods with average improvements of  $48.76\%$, $14.85\%$, and $11.45\%$ over the second-best method on old, regular, and new classes, respectively, and yields results comparable to the upper bound, \ie, the jointly trained model (Joint-T).
Throughout the whole incremental learning episodes, our approach not only achieves the least forgetting, but also yields the largest positive backward knowledge transfer (see the BWT column).
In most prior works, their BWT values are negative, indicating that learning new instruments only makes those previously learned ones be forgotten, rather than benefiting them.
In contrast, equipped with the hierarchical prompt parsing tree, our approach not only yields the positive BWT score, but also the highest one, demonstrating its great capability in alleviating catastrophic forgetting and refining past knowledge.
Meanwhile, with the help of the proposed self-reflection refining strategy, our approach achieves the highest FWT score (\ie, 6.96) with the DeepLabv3+ backbone.
It greatly outperforms other comparison approaches that adopt the same backbone as ours, indicating the superior performance of our approach in promoting new-instrument learning.

Similar improvements could also be found in laparoscopic cholecystectomy experiments in Table~\ref{tab2}.
Our model achieves the best performance across all metrics, including FWT and BWT, surpassing the state-of-the-art approach by 19.76\%, 18.46\%, and 6.16\% on old, regular, and new classes of surgical instruments, respectively.

We visualize the segmentation results of the proposed method in comparison with other SOTA methods, as shown in Fig.~\ref{vis}. 
Our method effectively learns new classes (\ie, the suction instrument in the third row), while maintaining strong performance on old (\ie, vessel sealer and grasping retractor in first row) and regular classes (\ie, prograsp forceps in the second row).
Additionally, we validate that our method remains robust and effective even when part-wise information is unavailable due to special circumstances.
The detailed strategy and experimental results are provided in the supplementary, along with additional discussions.

\subsection{Ablation Study}
\subsubsection{Analysis of the Key Components}
The effectiveness of each component in our framework is evaluated on the EndoVis 2017 and 2018 datasets (Table~\ref{tab3ablation}).
We consider the following settings, (a) using only the SAM backbone (1st-row), (b) introducing hierarchical prompt parsing tree (HPPT) upon the SAM backbone (2nd-row), (c) using hierarchical prompt parsing tree as well as the proposed self-reflection refining strategy (SRS) with the SAM backbone (3rd-row), (d) using only the DeepLabV3+ backbone (4th-row), and (e) applying the entire proposed method upon DeepLabV3+ backbone (last row).
Compared to using SAM alone, introducing HPPT improves average performance by 9.74\%.
Further incorporating the proposed SRS brings an overall performance gain of 1.55\%.
Our method is also compatible with other backbones. 
When applied to DeepLabV3+, it achieves a notable performance improvement of 13.30\%.
\subsubsection{Hyperparameter Studies}
\begin{figure}[t]
    \centering
    \includegraphics[width=0.46\textwidth]{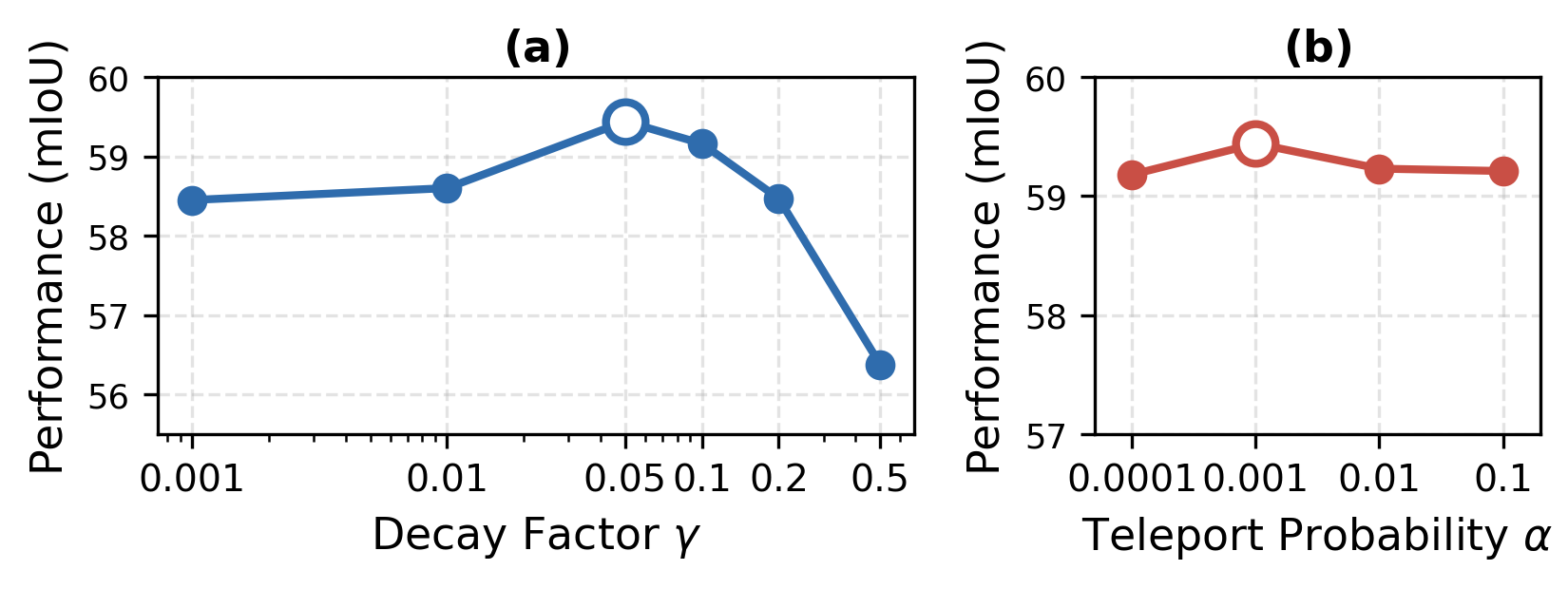}
    \caption{
    Effects of the decay factor $\gamma$ and the teleport probability $\alpha$ on model performance.
    }
  \label{fig:gamma-alpha}
\end{figure}
We investigate how the performance of the model is affected by different values of the decay factor $\gamma$, as shown in Figure~\ref{fig:gamma-alpha}~(a). 
A large decay factor may cause new knowledge to influence distant, irrelevant nodes, degrading performance.
In contrast, a small decay factor overly restricts the spread of new information, preventing it from reaching other nodes and hindering positive backward knowledge transfer.
The model achieves the best performance when $\gamma = 0.05$.
In addition, we also explore the impact of teleport probability $\alpha$ on model performance, as shown in Figure~\ref{fig:gamma-alpha} (b).
To maximally preserve the structural information of the hierarchical prompt parsing tree, we restrict $\alpha$ to small values approaching 0.
The performance peaks when $\alpha = 0.001$.
\subsubsection{Computational Cost Analysis}
We evaluate the computational efficiency of our method from multiple aspects, including the model size (Parameters), FLOPs, training time (Train.~time), and inference time (Infer.~time), as reported in Table~\ref{tab:complexity}. 
The second column (SAM) reports the results of using SAM alone for segmentation. 
The third column (Ours) reports results of our method built upon SAM as the backbone. 
The final column ($\Delta$ vs.\ SAM) quantifies the additional computational overhead introduced by our method compared to using SAM alone, showing that the overhead of our model is negligible.

\begin{table}[t]
  \centering
  \caption{Computational cost analysis.}
  \setlength{\tabcolsep}{10pt} %
  \renewcommand{\arraystretch}{1} %
  \resizebox{0.98\linewidth}{!}{
  \begin{tabular}{lccc}
    \toprule
    \textbf{Metric} & \textbf{SAM} & \textbf{Ours} & \textbf{$\Delta$ vs.\ SAM} \\
    \midrule
    Parameters (M)      & 636  & 639 & +0.47\% \\
    FLOPs (G)           & 2733  & 2751 & +0.65\% \\
    Train. time (s per iteration) & 0.901  & 0.915  & +1.55\% \\
    Infer. time (s per slice) & 0.464   & 0.466    & +0.43\% \\
    \bottomrule
  \end{tabular}
  }
  \label{tab:complexity}
\end{table}
   
\section{Conclusion}
We present a self-reflection hierarchical prompting framework that enables positive forward and backward knowledge transfer to improve the performance of class-incremental surgical instrument segmentation.
To promote positive forward knowledge transfer, we propose to structure instrument-aware prompts into a hierarchical prompt parsing tree that follows a general-to-specific knowledge hierarchy, to expose the reusable knowledge for new classes to facilitate their learning process.
Meanwhile, we realize positive backward knowledge transfer by proposing a self-reflection refining strategy to examine the knowledge association among early-acquired prompt partitions according to the topology structure of the tree, wisely refining historical knowledge without causing catastrophic forgetting.
Our framework is applicable to both CNN-based models and advanced transformer-based foundation models.
Experiments on two public surgical segmentation benchmarks have validated the effectiveness of our approach, yielding significant improvements over the state-of-the-art approaches.
\\
\\
\textbf{Acknowledgements: }The work described in this paper was supported in part by the Research Grants Council of the Hong Kong Special Administrative Region, China, under Project T45-401/22-N.
This work is also supported by the Hong Kong Research Grant Council under the Collaborative Research Fund C4042-23GF and General Research Fund 14214322, 14200623, 14203424, 14206325, and CUHK Strategic Seed Funding for Collaborative Research Scheme. The authors would also like to thank the support from Multi-scale Medical Robotics Center, AIR@InnoHK. 
{
    \small
    \bibliographystyle{ieeenat_fullname}
    \bibliography{0-ref}
}
\clearpage
\setcounter{page}{1}
\setcounter{section}{0}
\setcounter{figure}{0}
\setcounter{table}{0}
\section*{Supplementary}
In the supplementary, we additionally provide implementation details, training scheme descriptions, complete visualization comparisons, additional ablation studies, and further discussion regarding the limitations and future work.

\section{MLLM-based Part Count Estimation}
In most surgical video datasets, instrument annotations are provided in a part-wise manner, i.e., each part of the instrument is labeled as a different category.
Consequently, the number of parts for each instrument class can be directly obtained as a prior (all four datasets used in our paper follow this convention).
However, in real-world deployments, part-wise annotations may be unavailable, making it unclear how many parts an instrument consists of. 
To address this, we design a strategy that leverages an MLLM to automatically infer the number of parts when such annotations are missing (see Fig.~\ref{part}).
Concretely, we rely only on class-wise labels. Taking the Vessel Sealer as an example, we first perform a cutout based on the class label to isolate the instrument from all frames containing it, and feed these cropped instrument images into the MLLM with the prompt: “How many parts does this instrument consist of?” We then collect the answers derived by MLLM across samples. 
Finally, we apply majority voting to the responses and use the most frequent value as the estimated number of parts for each instrument.
We validated the proposed strategy on both the EndoVis2017~\cite{allan20192017} and EndoVis2018~\cite{allan20202018} datasets. 
As shown in Table~\ref{tab:endovis2017} and Table~\ref{tab:2018}, the results demonstrate that our method successfully assigns the correct part count to all instruments across both datasets.

\section{Implementation details}
We follow the preprocessing procedure described in \cite{chen2024ma} to preprocess the dataset.
Based on the supplementary materials, along with the provided code and the open-source implementations of \cite{chen2024ma} and \cite{tong2020digraph}, our project can be fully reproduced.
When using SAM, we adopt the SAM ViT-H~\cite{kirillov2023segment} version as our encoder $\theta_\text{Enc}$. 
In our practical implementation, for the prompt partition \( \text{P} \in \mathbb{R}^{u \times b} \), the number of tokens \( u \) is treated as a tunable hyperparameter.
Based on our grid search results, the best performance is achieved when $u$ is set to 12.
We employ the Adam optimizer with a learning rate of $1e^{-4}$ for the optimization. We use the PyTorch framework, and the model is trained with an NVIDIA A6000 GPU. 
\begin{figure}[t]
    \centering
    \caption{Illustration of HPPT construction.}
    \includegraphics[width=0.455\textwidth]{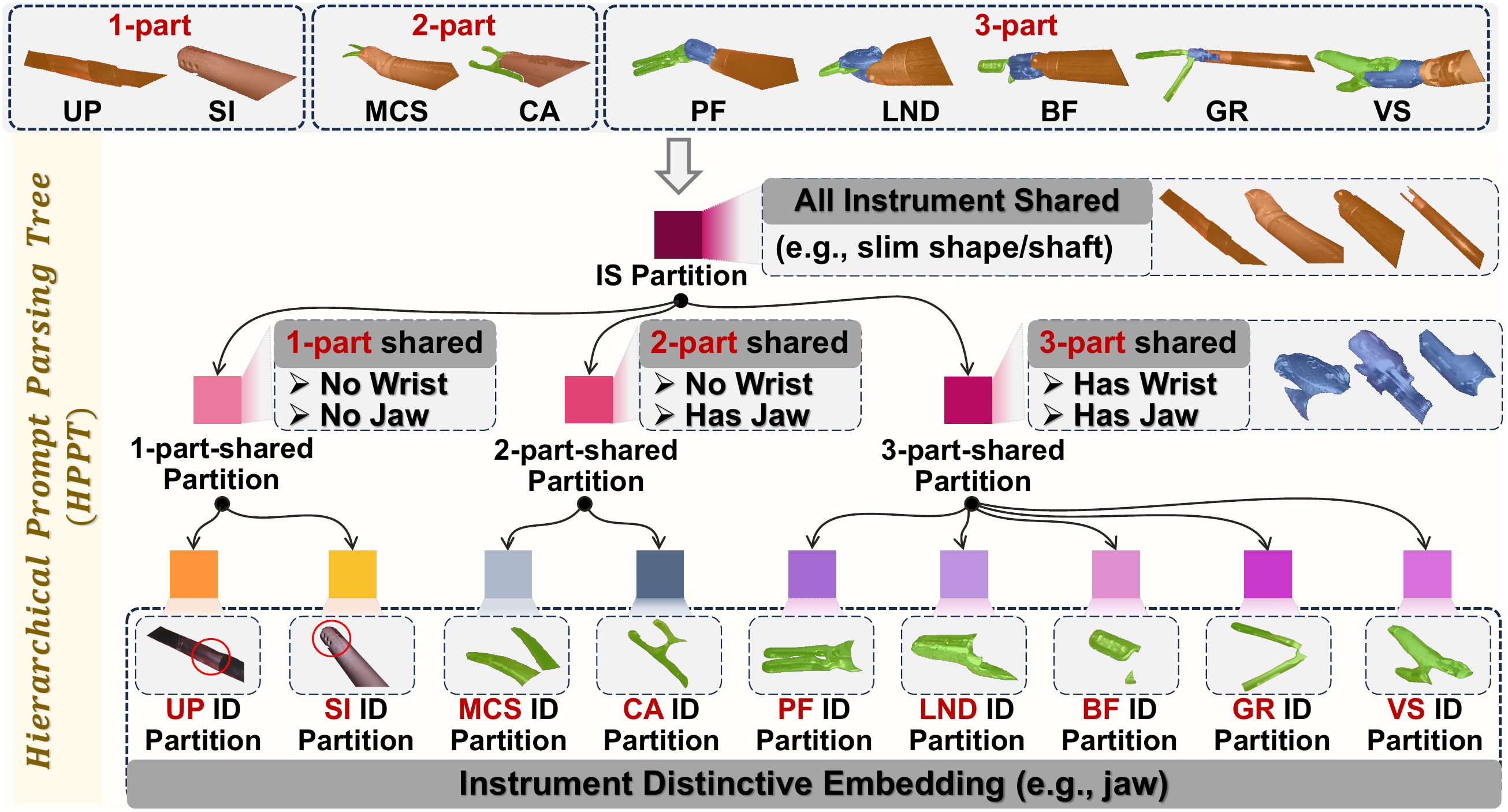}
  \label{fig:chart}
\end{figure}
\begin{figure*}[t]
  \centering
  \includegraphics[width=0.95\textwidth]{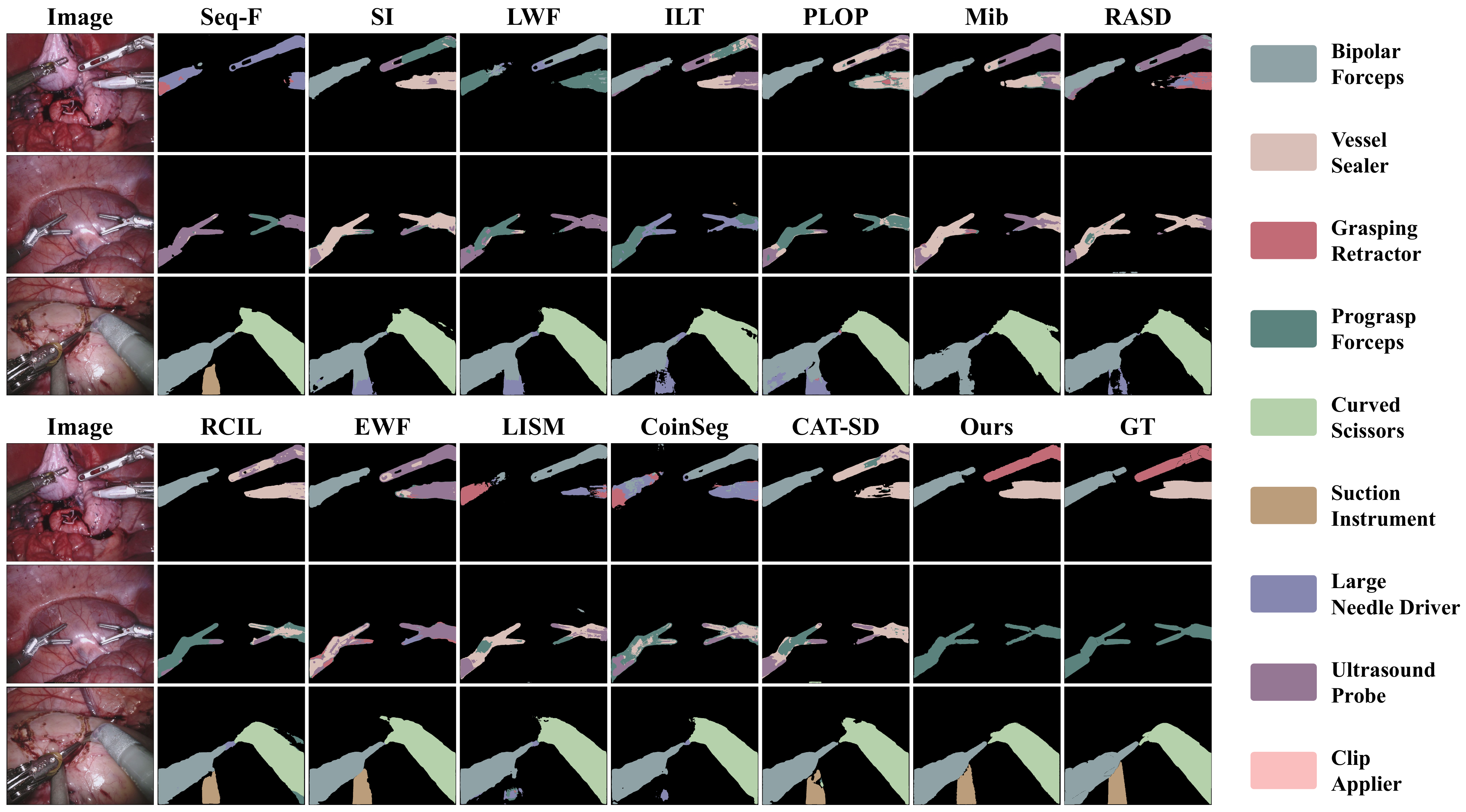}
    \caption{Visualization comparison of segmentation results across all methods. \textit{Rows 1–2, 4-5}: previous datasets. \textit{Row 3, 6}: current dataset.}
  \label{part}
\end{figure*}
\begin{figure*}[t]
  \centering
  \includegraphics[width=1\textwidth]{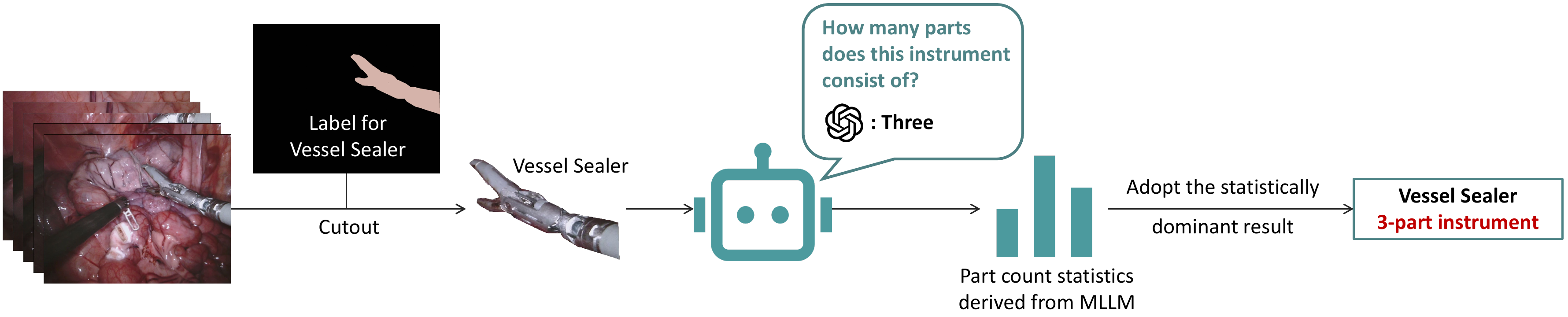}
  \caption{Automatic estimation of instrument part count.
  %
  %
  }
  \label{part}
\end{figure*}
\section{Metrics for Class-Incremental Learning}
To provide a deeper understanding of model performance in the class-incremental learning (CIL) setting, we adopt two widely-used evaluation metrics: Backward Transfer (BWT) and Forward Transfer (FWT)~\cite{apolinario2025code, diaz2018don, lopez2017gradient}. These metrics quantify how knowledge is transferred across tasks during the incremental learning process.
Due to the fundamental differences between class-incremental learning and other continual learning scenarios, we adopt specific BWT and FWT metrics to align with this setting.
\paragraph{Backward Transfer (BWT)}
BWT evaluates the influence of learning new data on the performance of previously learned (old and regular) classes. Specifically, for training episode $t$, we define BWT as the degradation (or improvement) in performance on old and regular classes compared to the previous episode $(t{-}1)$:
\begin{equation}
\text{BWT} = \frac{1}{|\mathcal{C}_t^{\text{old}} \cup \mathcal{C}_t^{\text{reg}}|} \sum_{c \in \mathcal{C}_t^{\text{old}} \cup \mathcal{C}_t^{\text{reg}}} \left( \text{IoU}_{t(c)} - \text{IoU}_{{t-1}(c)} \right),
\end{equation}
where $\text{IoU}_{t(c)}$ denotes the intersection over union (IoU) score for class $c$ after training at episode $t$, and $\text{IoU}_{{t-1}(c)}$ is the corresponding IoU score from the previous episode.\\
\textbf{Interpretation:}
\begin{itemize}
    \item $\text{BWT} < 0$: indicates \textit{forgetting}, meaning that the performance on previously learned (old and regular) classes has decreased after learning the current episode.

    \item $\text{BWT} > 0$: indicates \textit{positive backward transfer}, suggesting that learning new classes leads to improved performance on past classes. This may happen when the new data provides beneficial features that reinforce earlier knowledge.

    \item $\text{BWT} = 0$: indicates no change in past performance.
\end{itemize}

\paragraph{Forward Transfer (FWT)}  
FWT measures how prior learning benefits the model in learning new classes. For episode $t$, we define FWT as the difference between the IoU score of the model on newly introduced classes and the average IoU score obtained when training those classes in isolation (i.e., individual training):

\begin{equation}
\text{FWT} = \frac{1}{|\mathcal{C}_t^{\text{new}}|} \sum_{c \in \mathcal{C}_t^{\text{new}}} \left( \text{IoU}_{t(c)} - \text{IndIoU}_{(c)} \right)
\end{equation}
where $\text{IoU}_t(c)$ denotes the intersection over union score (IoU) for class $c$ after training at episode $t$, and $\text{IndIoU}(c)$ refers to the IoU score obtained when class $c$ is trained by individual training.\\
\textbf{Interpretation:}
\begin{itemize}

    \item $\text{FWT} < 0$: indicates \textit{negative forward transfer}, suggesting that prior knowledge interferes with the learning of new classes. 
    
    \item $\text{FWT} > 0$: indicates \textit{positive forward transfer}, meaning that knowledge from previous episodes facilitates learning of new classes. 
    
    \item $\text{FWT} = 0$: indicates no forward transfer.
\end{itemize}

\begin{table*}[t]
\centering
\caption{Part Count Inference Results on EndoVis2017 Dataset}
\renewcommand{\arraystretch}{1}

\begin{tabular}{
    >{\centering\arraybackslash}m{3.2cm}  
    >{\centering\arraybackslash}m{1.8cm}  
    >{\centering\arraybackslash}m{1.2cm}  
    >{\centering\arraybackslash}m{1.2cm}  
    >{\centering\arraybackslash}m{1.2cm}  
    >{\centering\arraybackslash}m{1.2cm}  
    >{\centering\arraybackslash}m{1.2cm}  
}
\toprule
\multirow{2}{*}{\textbf{Instrument}} &
\multirow{2}{*}{\textbf{Total Cutout}} &
\multicolumn{4}{c}{\textbf{Part Count Inferred by MLLM}} &
\multirow{2}{*}{\textbf{GT}} \\
\cmidrule(lr){3-6}
& & \textbf{1} & \textbf{2} & \textbf{3} & \textbf{Result} & \\
\midrule
Large Needle Driver        & 1197 & 156 & 213 & \underline{828} & 3 & 3 \\
Prograsp Forceps           & 1051 & 55  & 232 & \underline{764} & 3 & 3 \\
Bipolar Forceps            & 657  & 189 & 132 & \underline{336} & 3 & 3 \\
Ultrasound Probe           & 449  & \underline{294} & 103 & 52  & 1 & 1 \\
Curved Scissors  & 351  & 54  & \underline{205} & 92  & 2 & 2 \\
Vessel Sealer              & 386  & 122 & 53  & \underline{211} & 3 & 3 \\
Grasping Retractor         & 193  & 46  & 20  & \underline{127} & 3 & 3 \\
\bottomrule
\end{tabular}
\label{tab:endovis2017}
\end{table*}
\begin{table*}[t]
\centering
\caption{Part Count Inference Results on EndoVis2018 Dataset}

\renewcommand{\arraystretch}{1} %
\begin{tabular}{
    >{\centering\arraybackslash}m{3.2cm}  
    >{\centering\arraybackslash}m{1.8cm}  
    >{\centering\arraybackslash}m{1.2cm}  
    >{\centering\arraybackslash}m{1.2cm}  
    >{\centering\arraybackslash}m{1.2cm}  
    >{\centering\arraybackslash}m{1.2cm}  
    >{\centering\arraybackslash}m{1.2cm}  
}
\toprule
\multirow{2}{*}{\textbf{Instrument}} &
\multirow{2}{*}{\textbf{Total Cutout}} &
\multicolumn{4}{c}{\textbf{Part Count Inferred by MLLM}} &
\multirow{2}{*}{\textbf{GT}} \\
\cmidrule(lr){3-6}
& & \textbf{1} & \textbf{2} & \textbf{3} & \textbf{Result} & \\
\midrule
Large Needle Driver       & 278  & 58  & 96   & \underline{124}  & 3 & 3 \\
Prograsp Forceps          & 972  & 167 & 232  & \underline{573}  & 3 & 3 \\
Bipolar Forceps           & 1757 & 395 & 132  & \underline{1230} & 3 & 3 \\
Ultrasound Probe          & 156  & \underline{103} & 34   & 52   & 1 & 1 \\
Curved Scissors & 1554 & 205 & \underline{1257} & 92   & 2 & 2 \\
Suction Instrument        & 272  & \underline{159}  & 31   & 82  & 1 & 1 \\
Clip Applier              & 44   & 6  & \underline{26}    & 12   & 2 & 2 \\
\bottomrule
\end{tabular}
\label{tab:2018}
\end{table*}
\section{Training and Inference Scheme}
In the first episode, we freeze the encoder of SAM and jointly train the decoder, segmentation head, adapter, and the prompt parsing tree.
Since each image in the dataset typically contains more than one class of surgical instrument, we activate all instrument-aware prompts during the forward propagation, allowing each pixel to receive prediction scores for all current classes.
As a result, the model generates a set of binary segmentation masks, one for each class.
To fuse these predictions $\{\hat{y}_c\}_{c\in\mathcal{C}}$ into a final segmentation result $\hat{y} \in \mathcal{C}^{H \times W}$, we apply an argmax-based fusion. 
For each pixel location  $(i, j)$, the final predicted class is determined as:
\begin{equation}
\label{encode_feature}
\begin{aligned}
\hat{y}(i, j) = 
\begin{cases}
\displaystyle \arg\max_{c \in \mathbf{C}} \hat{y}_c(i, j), & \text{if } \displaystyle \max_{c \in \mathbf{C}} \hat{y}_c(i, j) > \tau \\
\text{background}, & \text{otherwise},
\end{cases}
\end{aligned}
\end{equation}
where $\tau$ denotes a confidence threshold used to ignore low-confidence predictions, and it is set to 0.5 based on our grid search.
We then apply the following objective function for optimization:
\begin{equation}
\label{t1object funtion}
\begin{aligned}
\min_{\substack{\mathcal{T}^t,\theta_a,\\\{\theta_{\text{seg}}^c\}_{c \in \mathbf{C}},\\\{\theta_\text{Dec}^n\}_{n=1}^N}}
\sum_{c\in \mathbf{C}} \mathcal{L}_{\text{ce}}\left( \theta_{\text{seg}}^c\left(\theta_\text{Dec}\left(f_\text{Enc}\left(x,\theta_a\right)\right), \mathcal{T}^t \right), y^c \right).
\end{aligned}
\end{equation}
This design ensures that, during backpropagation, the loss computed from the prediction and ground truth of a specific class is only propagated to the corresponding components.
As a result, each module learns knowledge that is specific and relevant to its associated class.
In the subsequent training process, each episode consists of two stages.
First, we enable the positive forward transfer. During this stage, the attention layers in the decoder, the segmentation heads for previously learned and regular classes, the adapter, and the prompt parsing tree are all frozen. Only the prompt partition corresponding to the new class is updated.
Once the new class has been trained, its prompt partition is inserted into the current prompt parsing tree.
Next, we conduct iterative self-reflection, where all modules—including the decoder, segmentation heads, adapter, and prompt parsing tree are frozen. In this stage, we train the $f_{dign}$ as introduced in the manuscript.
The output from the optimized $f_{dign}$ is then used as the updated prompt parsing tree for the next episode.
During inference, all instrument-aware prompts are retrieved and fused according to the aforementioned strategy to produce the final segmentation mask.
%
\section{Discussion}
Although prompt-based approaches and foundation models have recently emerged in surgical instrument segmentation, none have addressed the Class-Incremental Learning (CIL) setting. To the best of our knowledge, this work is the first to integrate prompt tuning with the Segment Anything Model (SAM) specifically for surgical instrument class incremental segmentation, thereby bridging a significant gap in the field.
The complexity of this task stems from the fine-grained nature of surgical scenes. Unlike natural domains that typically involve coarse-grained class shifts (e.g., distinguishing a bird from a tree), surgical CIL requires discriminating between instruments with highly similar metallic textures and slim structures, such as differentiating an L-hook from a J-hook.
Our method has been extensively validated on two prevalent surgeries, Nephrectomy (EndoVis 2017/2018) and Cholecystectomy (CholecSeg8k/M2CAI-Seg). In future work, we will extend our framework to additional procedures, such as prostatectomy.
\end{document}